\documentclass[letterpaper]{article} 
\usepackage{aaai2026}  
\usepackage{times}  
\usepackage{helvet}  
\usepackage{courier}  
\usepackage[hyphens]{url}  
\usepackage{graphicx} 
\usepackage{natbib}  
\usepackage{caption} 
\usepackage{algorithm}
\usepackage{algorithmic}
\usepackage{booktabs}
\usepackage{multirow}
\usepackage{amsmath}
\usepackage{amsfonts}
\usepackage{xcolor}

\usepackage{newfloat}
\usepackage{listings}
\DeclareCaptionStyle{ruled}{labelfont=normalfont,labelsep=colon,strut=off} 
\floatstyle{ruled}
\newfloat{listing}{tb}{lst}{}
\floatname{listing}{Listing}
\title{NeuSpring: Neural Spring Fields for Reconstruction and Simulation of Deformable Objects from Videos}
\author{
    Qingshan Xu\textsuperscript{\rm 1}\equalcontrib, Jiao Liu\textsuperscript{\rm 1}\equalcontrib, Shangshu Yu\textsuperscript{\rm 2}\thanks{Corresponding authors}, Yuxuan Wang\textsuperscript{\rm 1}, Yuan Zhou\textsuperscript{\rm 1\dag}, Junbao Zhou\textsuperscript{\rm 1}, \\
    Jiequan Cui\textsuperscript{\rm 3}, Yew-Soon Ong\textsuperscript{\rm 1,4}, Hanwang Zhang\textsuperscript{\rm 1}
}
\affiliations{
    \textsuperscript{\rm 1}Nanyang Technological University, 
    \textsuperscript{\rm 2}Northeastern University, 
    \textsuperscript{\rm 3}Hefei University of Technology, 
    \textsuperscript{\rm 4}A*STAR, Singapore \\

}

\usepackage{bibentry}

\begin{document}

\maketitle

\begin{abstract}
In this paper, we aim to create physical digital twins of deformable objects under interaction. Existing methods focus more on the physical learning of current state modeling, but generalize worse to future prediction. This is because existing methods ignore the intrinsic physical properties of deformable objects, resulting in the limited physical learning in the current state modeling. 
To address this, we present NeuSpring, a neural spring field for the reconstruction and simulation of deformable objects from videos. Built upon spring-mass models for realistic physical simulation, our method consists of two major innovations: 1) a piecewise topology solution that efficiently models multi-region spring connection topologies using zero-order optimization, which considers the material heterogeneity‌‌ of real-world objects. 2) a neural spring field that represents spring physical properties across different frames using a canonical coordinate-based neural network, which effectively leverages the spatial associativity of springs for physical learning. Experiments on real-world datasets demonstrate that our NeuSping achieves superior reconstruction and simulation performance for current state modeling and future prediction, with Chamfer distance improved by \textbf{20\%} and \textbf{25\%}, respectively.
\end{abstract}

\begin{links}
    \link{Code}{https://github.com/GhiXu/NeuSpring}
\end{links}

\section{Introduction}

\begin{figure*}
    \centering
    \includegraphics[width=\linewidth]{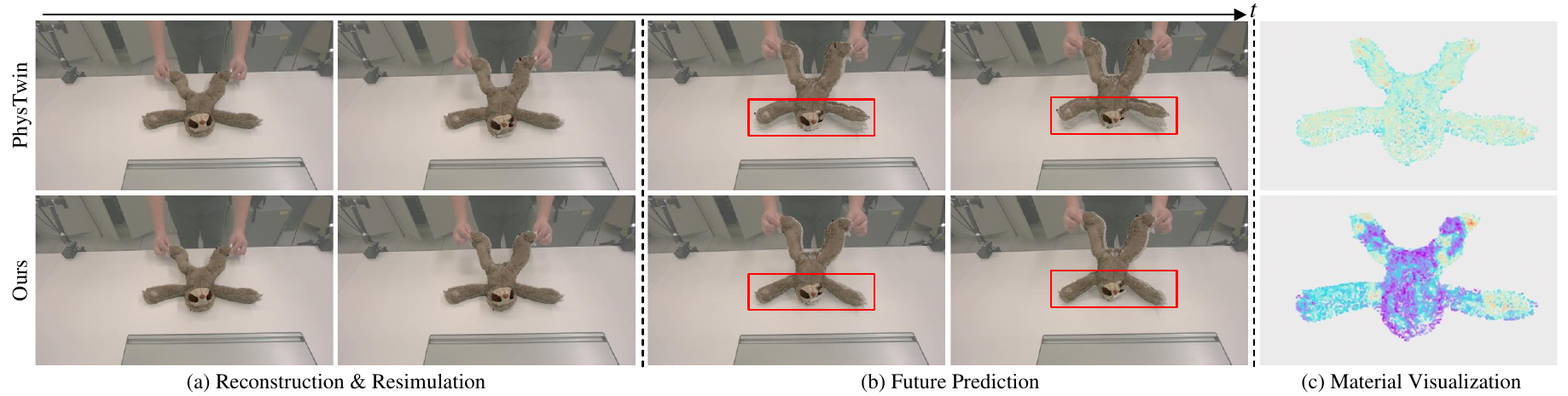}
    \caption{\textbf{Motivation of our NeuSpring.} We show the reconstruction \& resimulation and future prediction performance for PhysTwin~\cite{jiang2025phystwin} and our proposed method. We also visualize the approximated material based on the estimated physical property of springs--stiffness.}
    \label{fig:teaser}
\end{figure*}

Reconstruction and simulation of deformable objects is a fundamental problem in physical AI, with wide applications in robotics, animation, virtual reality, etc. The objective is to recover the geometry, appearance and physical properties of an object under interactions, allowing not only the reconstruction to closely match the input observations but also the future simulation to conform the realistic physical laws. This is critical for future prediction and planning. However, because of the complex physics involved in real-world deformable objects under interactions, it is difficult to recover the accurate physical properties of deformable objects, posing great challenges to the reconstruction and simulation of deformable objects.

Many works~\cite{park2021nerfies,kratimenos2024dynmf,yang2024deformable,wu20244d} have put effort into dynamic 3D reconstruction from videos. These methods focus on reconstructing the geometry, appearance and motion from input videos but ignore the underlying physics of deformable objects. This greatly limits their ability to predict future states. To model the intricate physics of deformable objects, there exist mainly two lines of research. Some neural dynamics methods~\cite{zhang2024dynamic,ma2023learning} adopt neural networks to model object dynamics, which take advantage of the powerful fitting capabilities of neural networks to learn complex physics. These pure learning-driven methods omit the basic physics modeling and require extensive training samples. Moreover, these methods are often limited to specific objects or motions. 

As another line of research, physics-driven methods~\cite{xie2024physgaussian,zhong2024reconstruction,jiang2025phystwin} build their physics models upon some powerful simulators, such as Material Point Method (MPM) and spring-mass models. 
These methods typically assume that objects exhibit certain regular physical properties, enabling the use of physical models—such as continuum mechanics or Newtonian mechanics—to describe their motion. By reconstructing the underlying properties and parameters of these models from video data, it becomes possible to reconstruct and predict the motion of deformable bodies. However, accurately identifying the appropriate parameters from sparse observations remains a significant challenge. A common issue is that, while such methods often align well with the observed motion in the input data, they tend to \textbf{generalize poorly} to future states or novel scenes, as illustrated in the top of Figure~\ref{fig:teaser}. 
This phenomenon can be attributed to: 
\begin{itemize}
    \item \emph{Independent learning:} existing physics-driven methods often adopt independent physics properties learning for each physical element, ignoring their spatial associativity and lacking structural regularization;
    \item \emph{Homogeneous assumption:} these methods often assume that the material of an object is homogeneous, resulting in a single material parameter or topology.
\end{itemize} 
This raises the key question we ask in this work: \emph{how to accurately learn the physics properties of deformable objects from the input observations?}  

To address this problem, based on spring-mass models, we introduce a novel implicit neural representation--\textbf{neural spring field} for reconstruction and simulation of deformable objects from videos, dubbed NeuSping. However, combining neural fields with spring-mass models to learn physics properties from videos is challenging due to: 1) spring-mass models contain non-differentiable parameters--spring connection topologies; 2) the spatial position of the spring varies across video frames. To this end, we introduce a two-stage learning strategy:
\begin{itemize}
    \item In the first stage, we present a piecewise topology solution. It divides deformable objects into different parts and models multi-region spring connection topologies under the assumption of homogeneous spring physics properties, with efficient zero-order optimizers. 
    \item In the second stage, we propose a canonical coordinate-based neural network to represent the physics properties of springs across different frames. We take as input the midpoint between spring endpoints in the canonical space, and estimate the spring physical properties through our carefully designed neural network.
\end{itemize} 
With the multi-region spring topologies and canonical spring physical properties, physical motions across frames are simulated and supervised by input trajectories to optimize spring properties. Integrated with the geometry and appearance modeled by 3D Gaussian splatting~\cite{kerbl20233d}, our proposed NeuSpring enables accurate reconstruction and simulation of deformable objects from videos.

Our approach inherits the benefits of neural methods and physics-driven methods: the expressive power of neural networks and the reduced requirement for training samples enabled by physical priors. Crucially, our method models the \textbf{material heterogeneity} of real-world objects and the \textbf{spatial associativity} of physics properties. As a result, our method can not only reconstruct dynamic 3D objects from input videos, but also generalize well to future prediction, as illustrated in the bottom of Figure~\ref{fig:teaser}. In particular, NeuSpring achieves \textbf{20\%} and \textbf{25\%} improvements in Chamfer distance for current state modeling and future prediction, respectively, demonstrating superior physical learning, see Table~\ref{tab:quant_indomain}. In summary, our main contributions are as follows:
\begin{itemize}
    \item We present a neural spring field for reconstruction and simulation of deformable objects from videos.
    \item We develop a canonical coordinate-based neural network to learn the physical properties of springs across different frames. It leverages the spatial continuity and coherence of springs to facilitate physical learning.
    \item We propose a piecewise topology solution to model multi-region spring connection topologies, which enables us to consider material heterogeneity.
\end{itemize}

\section{Related Work}

\noindent\textbf{Dynamic Scene Reconstruction.} 
Reconstructing dynamic scenes from videos has achieved significant advances in recent years~\cite{kratimenos2024dynmf,li2023dynibar,park2021nerfies,tretschk2021non}. Different from static scene reconstruction, dynamic scene reconstruction needs to consider the integration of motion modeling. Inspired by Neural Radiance Fields (NeRF)~\cite{mildenhall2021nerf}, early approaches~\cite{park2021nerfies,pumarola2021d} mainly focus on training a canonical NeRF and optimizing a deformable field to model dynamic scenes. Recently, 3D Gaussian Splatting (GS)~\cite{kerbl20233d} has emerged as a powerful technique due to its real-time efficiency and high-quality rendering, driving significant progress in dynamic scene reconstruction. Similarly, GS-based dynamic scene reconstruction methods~\cite{yang2024deformable,wu20244d} usually deform canonical 3D Gaussian primitives over time by optimizing deformation networks. Since these methods model motion information by conditioning the deformable field on timestep, they cannot capture real-world physics dynamics, preventing them from predicting future states and interactive modeling. 

\noindent\textbf{Neural Dynamics Modeling of Deformable Objects.} 
Several studies have explored the use of neural networks to approximate the dynamics of deformable objects~\cite{chen2022comphy,evans2022context,ma2023learning,wu2019learning,xu2019densephysnet}. A key advantage of these approaches lies in their ability to model complex state spaces and capture diverse physical properties. In particular, graph neural networks have been applied to simulate the dynamics of various materials, including plasticine~\cite{shi2023robocook,shi2024robocraft}, cloth~\cite{lin2022learning,pfaff2020learning}, fluids~\cite{li2018learning}, and stuffed animals~\cite{zhang2024dynamic}. More recently, physics-informed neural networks~\cite{raissi2019physics,wong2022learning,jin2024fourier,wong2021can} have also shown promise in modeling complex physical behaviors~\cite{xu2024precise}. However, as noted in \cite{jiang2025phystwin}, these methods often require large training data and are typically limited to a small number of predefined scenarios, which significantly constrains their generalizability.

\noindent\textbf{Physics-Driven Simulation of Deformable Objects.} 
Physical simulation is fundamental to physical AI~\cite{karniadakis2021physics,chiu2022can}, enabling physically plausible and expressive object deformations~\cite{NEURIPS2018_842424a1}. Classical models such as the mass-spring system~\cite{10.1145/2508363.2508406}, SPH~\cite{Monaghan_2005}, FEM~\cite{10.1145/142920.134016}, and MPM~\cite{10.1145/2897826.2927348} rely on prescanned or clean geometric data for realistic results.
Recent advances in 3D representations~\cite{mildenhall2021nerf,kerbl20233d,xu2024few,xu2024pushing} have integrated physics into geometric learning frameworks, including SDFs~\cite{xu2024precise,xu2025looks}, NeRF~\cite{Feng_2024_CVPR}, and Gaussians~\cite{xie2024physgaussian}. However, these approaches typically handle only small deformations under fixed forces, limiting real-world applicability.
The most related works, Spring-Gaussian~\cite{zhong2024reconstruction} and PhysTwin~\cite{jiang2025phystwin}, reconstruct deformable objects using mass-spring systems. Yet, Spring-Gaussian suffers from over-regularization, while PhysTwin’s separately learned parameters and homogeneity assumptions restrict physical accuracy. These limitations motivate our NeuSpring framework.

\section{Preliminary: Spring-Mass Model}

The spring-mass model is a classical framework for modeling the physical behavior of elastic and deformable objects. In this model, a deformable object is represented using a system of springs and masses, which can be formulated as a graph structure $\mathcal{G} = (\mathcal{V}, \mathcal{E})$, where $\mathcal{V}$ denotes the set of mass points defining the object's geometry, and $\mathcal{E}$ represents the set of springs connecting mass points. Both the mass points and springs are associated with physical parameters that characterize their behavior. Each mass point $i \in \mathcal{V}$ is associated with a position $\textbf{x}_i \in \mathbb{R}^{3}$ and a velocity $\textbf{v}_i \in \mathbb{R}^3$, which describe its state of motion at a given time step $t$. Each spring $e \in \mathcal{E}$ connects two mass points and is characterized by physical properties such as Young's modulus and damping ratio, which govern its elastic and dissipative behavior. This model provides a simple yet powerful basis for simulating the dynamics of deformable bodies.

In general, the spring-mass model simulates object deformation based on Newton's laws of motion. By computing the force acting on each mass point at every time step, the position and velocity of the mass point can be updated for the next time step. The force acting on the $i$-th mass point in $\mathcal{V}$ is determined by the combined effects of the neighboring mass points connected via springs:
\begin{equation}\label{Eq:Point_Force}
\begin{aligned}
\textbf{F}_i = \sum_{i,j \in \mathcal{E}} \left(\textbf{F}_{i,j}^\text{spring} + \textbf{F}_{i,j}^\text{dashpot} \right) + \textbf{F}_{i}^\text{ext},
\end{aligned}
\end{equation}
where the spring force $\textbf{F}_{i,j}^\text{spring}$ is calculated as: 
\begin{equation}\label{Eq:Spring_Force}
\begin{aligned}
\textbf{F}_{i,j}^\text{spring} = k_{ij} \left( ||\textbf{x}_j - \textbf{x}_i|| - l_{ij} \right) \frac{\textbf{x}_j - \textbf{x}_i}{||\textbf{x}_j - \textbf{x}_i||},
\end{aligned}
\end{equation}
the dashpot damping force $\textbf{F}_{i,j}^\text{dashpot}$ is calculated as:
\begin{equation}\label{Eq:Dashpot_Force}
\begin{aligned}
\textbf{F}_{i,j}^\text{dashpot} = - \gamma (\textbf{v}_i - \textbf{v}_j),
\end{aligned}
\end{equation}
and the external force $\textbf{F}_{i}^{\text{ext}}$ is incorporated to account for factors such as gravity, collisions, and user interactions.
In Eqs.~\eqref{Eq:Spring_Force} and \eqref{Eq:Dashpot_Force}, $k_{ij}$ denotes the stiffness of the spring connecting mass points $(i, j)$, $l_{ij}$ represents its rest length, and $\gamma$ is the damping coefficient associated with the dashpot. For collision handling—whether between mass points or between the object and external colliders—an impulse-based approach is employed.

The state of the spring system is updated using explicit Euler integration. Specifically, for all $i \in \mathcal{V}$, the velocities and positions at time step $t$ are updated as follows:
\begin{equation}\label{Eq:Euler}
\begin{aligned}
\textbf{v}_{i}^{t+1} = \delta \left( \textbf{v}_{i}^{t} + \Delta t  \cdot \frac{\textbf{F}_i}{m_i} \right), \\
\textbf{x}_{i}^{t+1} = \textbf{x}_{i}^{t} + \Delta t  \cdot \textbf{v}_{i}^{t+1},
\end{aligned}
\end{equation}
where $\delta$ is the drag damping.  
The simulation results of the spring-mass system are highly sensitive to physical parameters such as spring stiffness, damping, collision responses, and spring topology. This sensitivity becomes particularly pronounced with heterogeneous springs, where each spring may have distinct physical properties. As a result, the number of parameters grows substantially, posing significant challenges for parameter optimization in spring-mass-model-based dynamic 3D reconstruction.

\section{Problem Formulation}\label{sec:problem}

\begin{figure*}
    \centering
    \includegraphics[width=\linewidth]{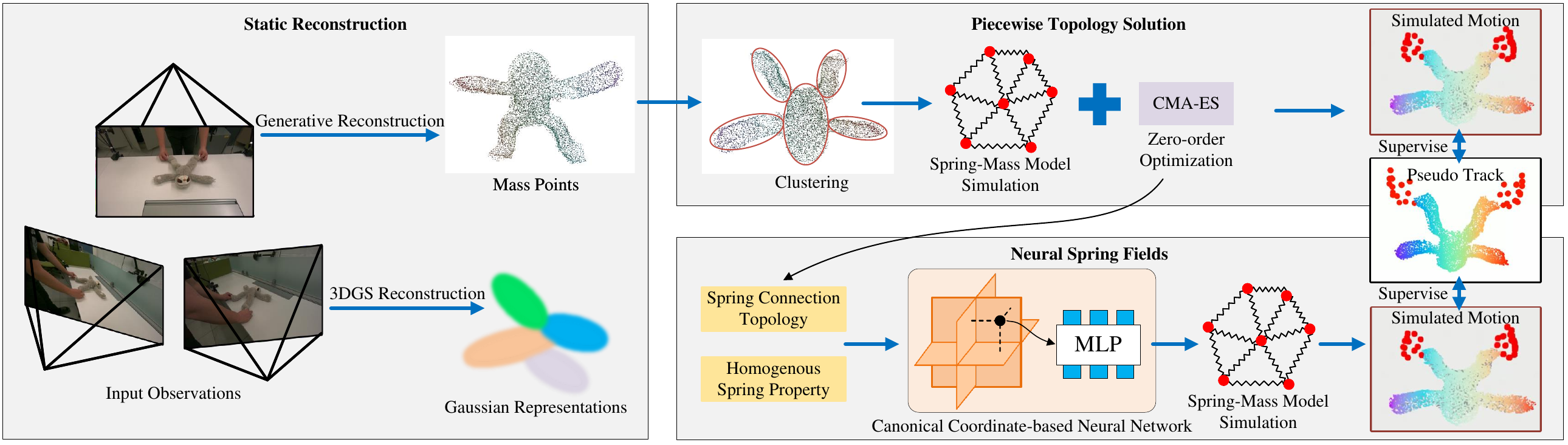}
    \caption{\textbf{Overview of Our NeuSpring.}}
    \label{fig:overview}
\end{figure*}

Our NeuSpring takes as input three RGBD videos of a deformable object under interaction to learn the appearance, geometry and physical properties for the deformable object. Inspired by \cite{jiang2025phystwin}, we formulate the dynamic reconstruction task as a two-stage optimization problem. 

In the first stage, we aim to enable the spring-mass model to accurately capture both the geometry and the deformation behavior of the target object. To this end, we jointly optimize the \textbf{physical property parameters} $\alpha$ (e.g., spring stiffness, damping coefficients, collision parameters, etc.) and the \textbf{topology} of the spring-mass model $\mathcal{G}$, allowing the physical system to support a faithful and expressive reconstruction of dynamic deformable objects. At each time frame $t$, the depth observation $\textbf{D}_t$ obtained from the camera is converted into a partial 3D point cloud $\textbf{X}_t$, which serves as a reference for adjusting the spring-mass system so that its simulation aligns with the observed data. Formally, the optimization problem for the first stage is defined as:
\begin{equation}\label{Eq:Phys_Opt}
    \begin{aligned}
    \min_{\alpha, \mathcal{G}} & \sum_{t} \left( \mathcal{L}_{\text{geometry}} (\textbf{X}_{t},\hat{\textbf{X}}_{t}) + \mathcal{L}_{\text{motion}} (\textbf{X}_{t},\hat{\textbf{X}}_{t}) \right) , \\
    s.t. & \ \hat{\textbf{X}}_{t+1} = f_{\alpha,\mathcal{G}}(\hat{\textbf{X}}_{t}, a_t), \\   
    \end{aligned}
\end{equation}
where $\mathcal{L}_{\text{geometry}}$ quantifies the single-direction Chamfer distance between the inferred state $\hat{\textbf{X}}_t$ and the partially observed point cloud $\textbf{X}_t$, encouraging the simulated geometry to align with the observed shape. Meanwhile, $\mathcal{L}_{\text{motion}}$ measures the motion tracking error between the predicted point $\hat{\textbf{x}}_i^t$ and its corresponding observed trajectory point $\textbf{x}_i^t$, enforcing temporal consistency of the simulated motion. 
The state of the spring-mass system, i.e., the velocity and position of the mass points, is updated via $f_{\alpha,\mathcal{G}}$ by using the explicit Euler integration. 
Following \cite{jiang2025phystwin}, the observed trajectories are extracted using the vision foundation model CoTracker3~\cite{karaev2410cotracker3}, and then lifted into 3D space via depth map unprojection.

In the second stage, since the spring-mass model captures only physical states without rendering, we establish a mapping from mass points to image space to bridge physics and observation. This ensures alignment between the simulated object's \textbf{appearance} and the observed frames for accurate dynamic reconstruction. 
Let $g_{\phi}$ denote the mapping function from the mass points to the rendered image, where $\phi$ represents the learnable parameters. The goal of the optimization in this stage is to minimize the discrepancy between the rendered predictions and the corresponding ground-truth observations. Accordingly, the overall optimization problem can be formulated as:
\begin{equation}\label{Eq:Render_Opt}
    \min_{\phi}  \sum_{t,c} \mathcal{L}_{\text{render}} (\textbf{I}_{t,c},\hat{\textbf{I}}_{t,c}), \
    s.t.  \ \hat{\textbf{I}}_{t,c} = g_{\phi}(\hat{\textbf{X}}_{t}, c), \\   
\end{equation}
where $c$ denotes the camera index, $\textbf{I}_{t,c}$ and $\hat{\textbf{I}}_{t,c}$ are the ground truth and rendered images from view $c$ at time $t$, respectively. The loss term $\mathcal{L}_{\text{render}}$ combines the pixel-wise $L_1$ loss with a D-SSIM component to encourage both photometric accuracy and structural consistency between the rendered and ground truth images.

In physics-driven dynamic reconstruction, physical simulation serves as the backbone for modeling the dynamic deformation of an object. The objective of this work is to improve the simulation fidelity by introducing NeuSpring, a more accurate and adaptable physical model for complex deformations. Consequently, our focus lies on the first-stage optimization, while the second stage follows the method in \cite{jiang2025phystwin} without modification. The following sections detail the proposed NeuSpring framework.

\section{Method}

Figure~\ref{fig:overview} illustrates the pipeline of NeuSpring. Given input sparse RGB inputs, we first utilize the 3D generative model TRELLIS~\cite{xiang2025structured} to generate mass points. Then, we adopt our piecewise topology solution to build spring topologies and initialize the physical properties of springs. Next, these outcomes are input to our neural spring fields proposed to further optimize physical properties. In this way, the dynamics of the mass points is obtained to guide the dynamic rendering of Gaussian representations.

\subsection{Neural Spring Fields}

As analyzed above, the core of reconstruction and simulation of deformable objects is to estimate accurate physical properties for deformable objects, which capture real-world physics dynamics. Previous spring-mass-based dynamic reconstruction methods~\cite{jiang2025phystwin,zhong2024reconstruction} typically adopt independent learning to optimize the physical property parameters of each spring. This ignores the spatial associativity of spring properties, leading to the limited physics learning in the current state modeling. To address this problem, based on spring-mass models, we introduce a neural spring field to learn the physical properties of springs from input observations. 

Our neural spring field is conditioned on the spatial information of springs so as to consider the spatial associativity of physical properties. Specifically, our neural spring field $\mathcal{S}$ is defined in the canonical space to represent the physical properties of springs across different frames. For each spring $e \in \mathcal{E}$ that connects two mass points $(i,j)$, we compute the midpoint between these points to represent the spatial information of each spring:
\begin{equation}
    \textbf{x}_\text{mid}(e) = \frac{(\textbf{x}_i + \textbf{x}_j)}{2}.
\end{equation}
Then, we use a neural network $F_\theta$ to learn the physical properties of each spring by querying its midpoint coordinate:
\begin{equation}
    \mathcal{S}(e) = F_\theta(\textbf{x}_\text{mid}(e)),
\end{equation}
where $\theta$ represents learnable parameters. More concretely, the $F_\theta$ adopts a tri-plane representation~\cite{chan2022efficient} which is a hybrid explicit-implicit network architecture and is efficient for evaluation. It contains three 2D feature planes $(\textbf{f}_{xy}, \textbf{f}_{yz}, \textbf{f}_{xz})$ with a spatial resolution of
$N \times N$ and $C$ feature channels each and a shallow Multi-Layer Perceptron (MLP). To learn $\mathcal{S}(e)$ at position $\textbf{x}_\text{mid}(e)$, we project $\textbf{x}_\text{mid}(e)$ onto each of the three planes and query the corresponding features on each plane by bilinear interpolation. By aggregating these features through the sum operation and passing them into the MLP with the Fourier feature transform, we obtain $\mathcal{S}(e)$.

\subsection{Piecewise Topology Solution}

In the spring field, the input is defined as the midpoints between pairs of connected points, which requires prior knowledge of the spring topology. Since this topological information is not encoded inherently in the neural spring field itself, it must be determined before training the neural spring field. Moreover, the spring topology significantly impacts simulation behaviors and thus affects dynamic reconstruction accuracy. In previous work~\cite{jiang2025phystwin,zhong2024reconstruction}, K-nearest neighbor (KNN) algorithms are commonly employed to construct topological structures by connecting nearby mass points with springs. The choice of KNN hyperparameters determines the final topology.  

While KNN provides an effective mechanism for defining connectivity, using a uniform set of KNN hyperparameters across the entire system results in a homogeneous spring connection density. However, real-world objects often exhibit heterogeneous material properties, and such uniform connection schemes may fail to capture this variability. 

To address this limitation, we propose a piecewise topology construction method for the spring-mass system. Specifically, we first cluster the mass points in the spring system. After clustering, we assign independent KNN hyperparameters—namely, the maximum number of neighbors and the search radius—to each cluster. This allows different regions of the object to have varying spring connection densities, thereby better reflecting material heterogeneity. To determine the optimal KNN hyperparameters for each cluster, we treat these hyperparameters as decision variables and solve the optimization problem defined in Eq.~\eqref{Eq:Phys_Opt}. Since the KNN hyperparameters are non-differentiable, we adopt a gradient-free optimization approach. In particular, we utilize the Covariance Matrix Adaptation Evolution Strategy (CMA-ES), a widely used evolutionary algorithm, to solve this problem. 

It is important to note that evaluating the objective function in Eq.~\eqref{Eq:Phys_Opt} requires not only the spring system topology but also the physical parameters of the springs. Therefore, during the optimization process, we assume homogeneous spring properties and include these physical parameters as part of the decision variables. The resulting optimized topology and homogeneous physical parameters are then used as the initialization for training the neural spring field. 
In this way, our neural spring field is modified as:
\begin{equation}
    \mathcal{S}(e) = \mathcal{S}_0 + F_\theta(\textbf{x}_\text{mid}(e)),
\end{equation}
where $\mathcal{S}_0$ is the obtained physical parameters in this phase, which provides a good initialization for our neural spring field learning.

\subsection{Reconstruction and Simulation with NeuSpring}

With our proposed NeuSpring, the physical properties parameters are learned to support the dynamic rendering of deformable objects. Specifically, following the practices in \cite{jiang2025phystwin,zhong2024reconstruction}, we adopt 3D Gaussian Splatting~\cite{kerbl20233d} to compute parameterized Gaussian points in the canonical frame, which models the static appearance and geometry of deformable objects. For the appearance and geometry in other frames, we leverage Linear Blend Skinning (LBS) to interpolate the motions of Gaussian points based on the motions of neighboring mass points~\cite{zhang2024dynamic,jiang2025phystwin}. By splatting these moved Gaussian points and implementing $\alpha$-blending,  
we efficiently render deformable objects across frames for realistic reconstruction and simulation.

\section{Experiments}

\begin{figure*}[t]
    \centering
    \includegraphics[width=0.98\linewidth]{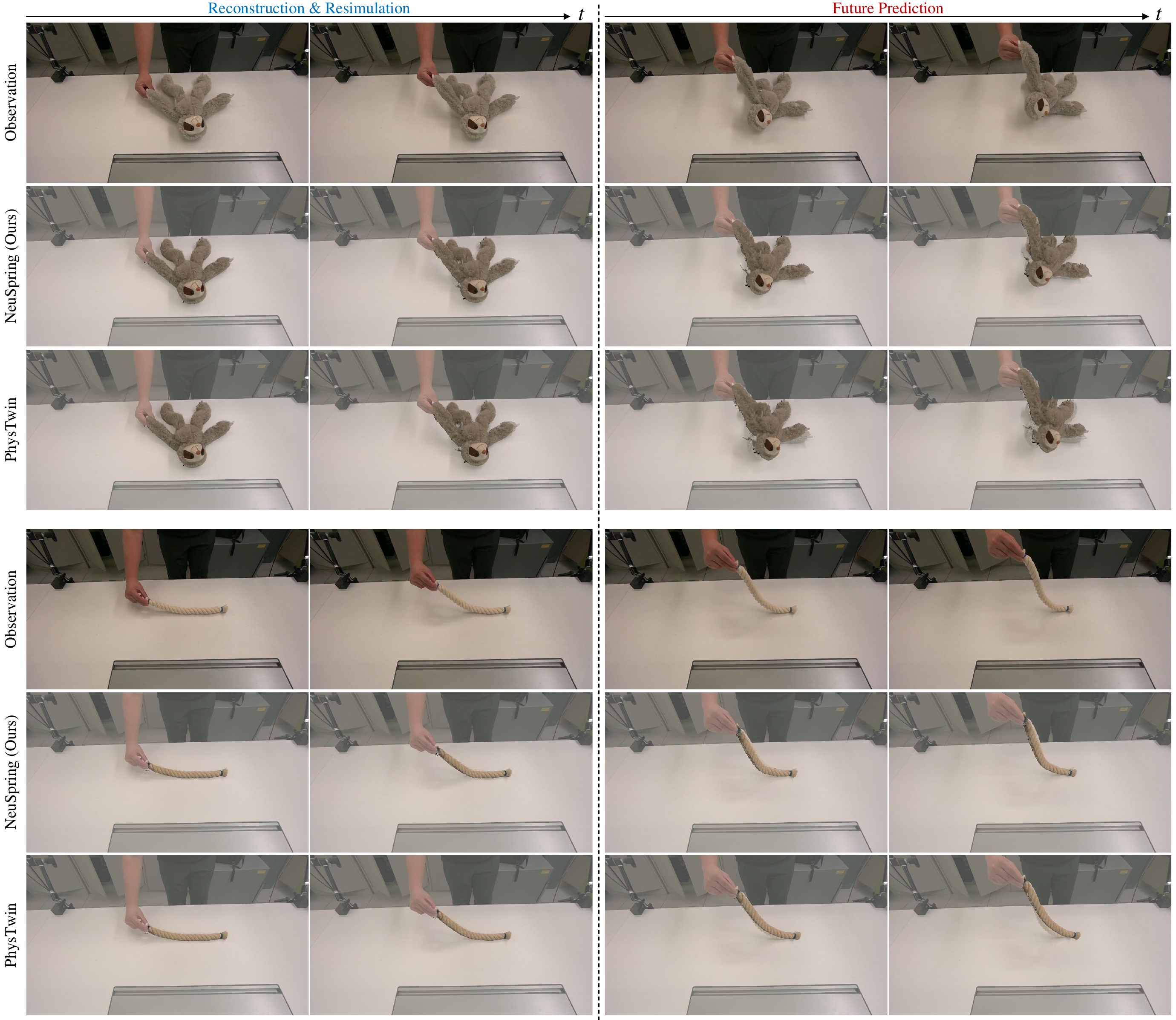}
    \caption{\textbf{Qualitative Results on Reconstruction \& Resimulation and Future Prediction.}}
    \label{fig:recon_simu}
\end{figure*}

\begin{table*}[ht]
\centering
\small
\setlength{\tabcolsep}{0.95mm} 
{
\begin{tabular}{@{} l rr rr rr rr rr rr @{}}
\toprule
Task &  \multicolumn{6}{c}{Reconstruction \& Resimulation} & \multicolumn{6}{c}{Future Prediction} \\
\cmidrule(lr){2-7} \cmidrule(lr){8-13}
Method & CD $\downarrow$ & Track Error $\downarrow$ & IoU \% $\uparrow$ & PSNR $\uparrow$ & SSIM $\uparrow$ & LPIPS $\downarrow$ &  CD $\downarrow$ & Track Error $\downarrow$ & IoU \% $\uparrow$ & PSNR $\uparrow$ & SSIM $\uparrow$ & LPIPS $\downarrow$ \\
\midrule
Spring-Gaus & 0.041 & 0.050 & 57.6 & 23.445 & 0.928 & 0.102 & 0.062 & 0.094 & 46.4 & 22.488 & 0.924 & 0.113 \\
GS-Dynamics & 0.014 & 0.022 & 72.1 & 26.260 & 0.940 & 0.052 & 0.041 & 0.070 & 49.8 & 22.540 & 0.924 & 0.097 \\
PhysTwin & \underline{0.005} & \underline{0.009} & \underline{84.4} & \underline{28.214} & \underline{0.945} & \underline{0.034} & \underline{0.012} & \underline{0.022} & \underline{72.5} & \underline{25.617} & \underline{0.941} & \underline{0.055} \\
NeuSpring (\textbf{Ours}) & \textbf{0.004} & \textbf{0.008} & \textbf{86.2} & \textbf{28.664} & \textbf{0.947} & \textbf{0.031} & \textbf{0.009} & \textbf{0.017} & \textbf{75.9} & \textbf{26.361} & \textbf{0.944} & \textbf{0.049} \\
\bottomrule
\end{tabular}
}
\caption{
    \textbf{Quantitative Results on Reconstruction \& Resimulation and Future Prediction.} 
    Our NeuSpring framework consistently outperforms the baselines across all metrics.
    }
\label{tab:quant_indomain}
\vspace{-0.2cm}
\end{table*}

\begin{table*}[ht]
\centering
\small
\setlength{\tabcolsep}{0.35mm} 
{
\begin{tabular}{@{} l cc rr rr rr rr rr rr @{}}
\toprule
Task & & & \multicolumn{6}{c}{Reconstruction \& Resimulation} & \multicolumn{6}{c}{Future Prediction} \\
\cmidrule(lr){4-9} \cmidrule(lr){10-15}
Method & PT & NSF & CD $\downarrow$ & Track Error $\downarrow$ & IoU \% $\uparrow$ & PSNR $\uparrow$ & SSIM $\uparrow$ & LPIPS $\downarrow$ &  CD $\downarrow$ & Track Error $\downarrow$ & IoU \% $\uparrow$ & PSNR $\uparrow$ & SSIM $\uparrow$ & LPIPS $\downarrow$ \\
\midrule
Baseline & & & 0.0054 & 0.0095 & 83.5 & 28.033 & 0.944 & 0.035 & 0.0118 & 0.0230 & 70.2 & 25.281 & 0.939 & 0.058 \\
Model-A & \checkmark & & 0.0050 & 0.0091 & 84.8 & 28.278 & 0.945 & 0.033 & 0.0114 & 0.0223 & 72.3 & 25.720 & 0.941 & 0.054 \\
Model-B & & \checkmark & 0.0048 & 0.0085 & 83.9 & 28.114 & 0.945 & 0.034 & 0.0101 & 0.0205 & 71.6 & 25.638 & 0.940 & 0.056 \\
NeuSpring (\textbf{Ours}) & \checkmark & \checkmark & \textbf{0.0043} & \textbf{0.0077} & \textbf{86.2} & \textbf{28.664} & \textbf{0.947} & \textbf{0.031} & \textbf{0.0087} & \textbf{0.0175} & \textbf{75.9} & \textbf{26.361} & \textbf{0.944} & \textbf{0.049} \\
\bottomrule
\end{tabular}
}
\caption{
    \textbf{Ablations of Our NeuSpring.} PT: Piecewise Topology solution. NSF: Neural Spring Fields. 
}
\label{tab:quant_ablation}
\end{table*}

\subsection{Experimental Settings}

\noindent\textbf{Dataset.} We evaluate NeuSpring on the dataset collected by PhysTwin \cite{jiang2025phystwin}, which contains 22 scenarios. Each scenario provides three RGBD videos capturing human interactions with various deformable objects with different physical properties. These scenarios can be categorized into six types of deformable objects, including cloth, ropes, packages, stuffed zebra, stuffed sloth and stuffed dinosaur. For each scenario, the RGBD videos are split into a training set and a testing set following a 7:3 ratio. The training set is used to train our NeuSpring for reconstruction and resimulation, while the test set is used to evaluate its generalization ability, i.e., future prediction.

\noindent\textbf{Baselines.} We compare our NeuSpring with state-of-the-art methods, including GS-Dynamics~\cite{zhang2024dynamic}, Spring-Gaus~\cite{zhong2024reconstruction} and PhysTwin~\cite{jiang2025phystwin}. The first one is a neural dynamics method while the last two are physics-driven methods.

\noindent\textbf{Evaluation Metrics.} To evaluate how our predictions match the observations, we adopt Chamfer Distance (CD), tracking error and image assessment metrics. The image assessment metrics include Peak Signal-to-Noise Ratio (PSNR), Structural Similarity Index Measure (SSIM)~\cite{wang2004image}, Learned Perceptual Image Patch Similarity (LPIPS)~\cite{zhang2018unreasonable} and silhouette alignment using IoU.

\noindent\textbf{Implementation Details.} Our implementation is based on PyTorch framework~\cite{paszke2019pytorch} and the simulation is built on the warp implementation of spring-mass models~\cite{macklin2022warp}. In the piecewise topology solution stage, we empirically select five clusters to represent flexible topologies of deformable objects with agglomerative clustering. In terms of our canonical coordinate-based neural network architecture, the spatial resolution of tri-plane representation is adaptively set to $N=0.85\cdot\sqrt{|\mathcal{E}|}$, and the feature channel is set to $C=32$. We adopt a 3-layer MLP with 128 hidden units. 
All experiments are performed on one NVIDIA RTX 3090 GPU.

\subsection{Comparison}

\noindent\textbf{Reconstruction \& Resimulation.} 
We report the reconstruction \& resimulation performance of different methods in the left column of Table~\ref{tab:quant_indomain}. We observe that our method achieves the best performance across all metrics.  
In particular, NeuSpring outperforms PhysTwin by \textbf{20.0\%}, \textbf{11.1\%}, \textbf{2.1\%} and \textbf{8.8\%} in CD, tracking error, IoU and LPIPS, respectively.  
As shown in Eq.~\eqref{Eq:Phys_Opt}, the CD and tracking error metrics reflect the accuracy of physical learning, our noticeable improvements in these two metrics demonstrate that NeuSpring can better learn the physical properties of deformable objects. Moreover, our better physical learning further boosts the performance of dynamic rendering.   
The qualitative results in Figure~\ref{fig:recon_simu} further validate the effectiveness of our proposed NeuSpring.

\noindent\textbf{Future Prediction.} The right column of Table~\ref{tab:quant_indomain} further shows the future prediction capability of different methods, which is an important component for digital twins. As can be seen, our method outperforms baseline methods across all metrics. In particular, NeuSpring outperforms PhhysTwin by \textbf{25.0\%}, \textbf{22.7\%}, \textbf{4.5\%} and \textbf{10.9\%} in CD, tracking error, IoU and LPIPS, respectively. We show qualitative visualizations of the future prediction of different methods in Figure~\ref{fig:recon_simu}. We note that although PhysTwin can achieve competitive reconstruction and resimulation, it cannot align its future predictions well with the observations in unseen frames. This demonstrates that the learning of PhysTwin is limited to the seen frames. In contrast,  NeuSpring can not only reconstruct seen frames well, but also predict unseen frames precisely. This provides strong evidence that our learned physics dynamics is more accurate than the baseline methods.

\subsection{Analysis}

\noindent\textbf{Ablation Study.} To evaluate the effect of our proposed contributions, we adopt PhysTwin as our baseline model to carry out ablation studies. Note that, we rerun the baseline model in our machine. Then, we progressively replace the corresponding modules in PhysTwin by our designs to investigate their efficacy. The results are reported in Table~\ref{tab:quant_ablation}.

By adding our designed piecewise topology solution to the baseline, all metrics are improved. This shows that this design better considers the material heterogeneity of deformable objects, providing more flexibility for topology optimization. Therefore, it facilitates physical learning. 
On the other hand, our proposed neural spring fields enhance both the baseline and Model-A. This strategy introduces spatial associativity for springs, serving as a structural regularizer for physical learning. Therefore, the improvements in the CD and tracking error metrics are more pronounced. 

\begin{figure}[t]
    \centering
    \includegraphics[width=\linewidth]{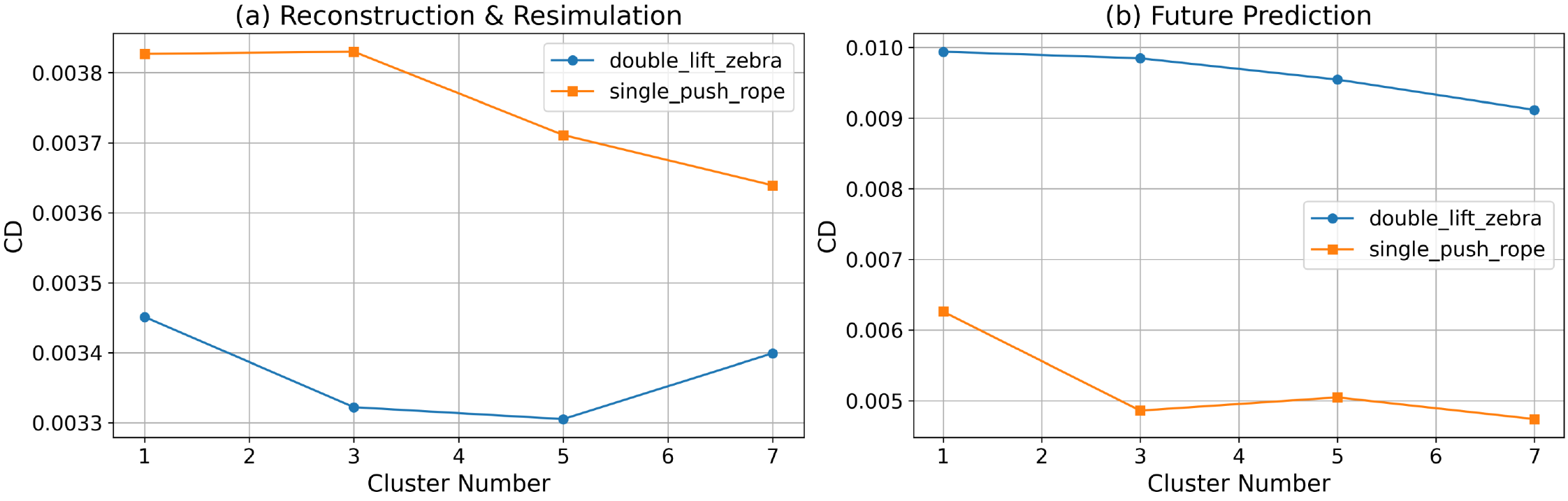}
    \caption{\textbf{Effect of the Cluster Number.}}
    \label{fig:cluster}
\end{figure}

\noindent\textbf{Effect of the Cluster Number.} Figure~\ref{fig:cluster} shows the effect of the cluster number on two scenarios. As the number of clusters increases, the CD metrics of reconstruction \& resimulation and future prediction almost decrease. This further demonstrates that equipping the spring-mass model with piecewise topologies is beneficial to physical learning. Generally, more clusters will allow the spring-mass model to consider more topology structures, providing more flexibility. However, this increases the number of parameters to be optimized, thus exacerbating the complexity of zeroth-order optimization. As a result, it typically requires more function evaluations and exhibits slower convergence. Thus, we empirically set the cluster number to five for all scenarios.    
 
\noindent\textbf{Effect of Network Design.} To investigate how our designed network learns the spatial associativity of springs, we conduct two experiments: removing the tri-plane representation (TP) and replacing adaptive resolution (AR) with fixed resolution $N=256$. The results are reported in Table~\ref{tab:effect_net}. We observe that, without explicit spatial modeling, the pure MLP cannot effectively learn the spatial associativity for springs,  obviously degrading the performance. With AR, our method can better consider the shape of deformable objects to model the spatial associativity, further improving the performance. 

\begin{table}[t]
\centering
\scriptsize
\setlength{\tabcolsep}{0.9mm} 
{
\begin{tabular}{@{} l rr rr rr rr @{}}
\toprule
Task &  \multicolumn{4}{c}{Reconstruction \& Resimulation} & \multicolumn{4}{c}{Future Prediction} \\
\cmidrule(lr){2-5} \cmidrule(lr){6-9}
Method & CD $\downarrow$ & TE $\downarrow$ & IoU \% $\uparrow$ & LPIPS $\downarrow$ &  CD $\downarrow$ & TE $\downarrow$ & IoU \% $\uparrow$ & LPIPS $\downarrow$ \\
\midrule
w/o TP & 0.0052 & 0.0091 & 84.1 & 0.035 & 0.0112 & 0.0216 & 72.1 & 0.056 \\
w/o AR & \underline{0.0045} & \underline{0.0083} & \underline{85.7} & \underline{0.032} & \underline{0.0095} & \underline{0.0186} & \underline{74.9} & \underline{0.052} \\
Ours& \textbf{0.0043} & \textbf{0.0077} & \textbf{86.2} & \textbf{0.031} & \textbf{0.0087} & \textbf{0.0175} & \textbf{75.9} & \textbf{0.049} \\
\bottomrule
\end{tabular}
}
\caption{
    \textbf{Effect of Network Design.} TP: Tri-Plane. AR: Adaptive Resolution of TP. TE: Track Error.}
\label{tab:effect_net}
\end{table}

\section{Conclusion}

In this paper, we propose Neural Spring fields, NeuSpring, for reconstruction and simulation of deformable objects from videos. We design the piecewise topology solution and neural spring field learning to consider the material heterogeneity and spatial associativity for spring-mass models, respectively. This facilitates the physical learning of deformable objects, enabling more accurate reconstruction and simulation. Extensive experiments demonstrate that NeuSpring achieves state-of-the-art performance in current state modeling and future prediction, offering valuable insights to construct effective physical digital twin for deformable objects from sparse videos.   

\bibliography{aaai2026}

\end{document}